\documentclass{article}

\usepackage[T1]{fontenc}
\usepackage[utf8]{inputenc}
\usepackage{microtype}
\usepackage{amsmath,amssymb,amsthm,mathtools,bm}
\usepackage{booktabs,tabularx,array,multirow}
\usepackage{siunitx}
\usepackage{graphicx}
\usepackage{enumitem}
\usepackage{algorithm}
\usepackage{algorithmic}

\usepackage{xcolor}
\usepackage[colorlinks=true,allcolors=blue]{hyperref}
\hypersetup{
  pdftitle={Certified Task-Conditioned Active Observability},
  pdfauthor={Linzhe Zhang, Changming Xu}
}
\usepackage[nameinlink,noabbrev]{cleveref}
\usepackage{placeins}
\usepackage[preprint]{icml2026}

\graphicspath{{figures/}}
\newtheorem{theorem}{Theorem}[section]

\newtheorem{corollary}[theorem]{Corollary}
\theoremstyle{definition}
\newtheorem{definition}[theorem]{Definition}

\theoremstyle{remark}

\newcommand{\AO}{\mathsf{AO}}
\newcommand{\DT}{\mathsf{D}}
\newcommand{\KL}{D_{\mathrm{KL}}}
\newcommand{\Hset}{\mathcal{H}}
\newcommand{\Aset}{\mathcal{A}}
\newcommand{\Yset}{\mathcal{Y}}
\newcommand{\Tset}{\Theta}
\newcommand{\Fhist}{\mathcal{F}}
\newcommand{\ind}{\mathbf{1}}

\icmltitlerunning{Certified Task-Conditioned Active Observability}

\begin{document}

\twocolumn[
\icmltitle{Certified Task-Conditioned Active Observability}
\begin{icmlauthorlist}
  \icmlauthor{Linzhe Zhang}{neu}
  \icmlauthor{Changming Xu}{neu}
\end{icmlauthorlist}
\icmlaffiliation{neu}{Graduate School, Northeastern University}
\icmlcorrespondingauthor{Linzhe Zhang}{cfmy007@gmail.com}
\icmlcorrespondingauthor{Changming Xu}{changmingxu@neuq.edu.cn}
\icmlkeywords{Active Observability, State Estimation, Sequential Decision Making, Representation Learning}
\vskip 0.3in
]
\printAffiliationsAndNotice{}

\begin{abstract}
Before an autonomous agent can act upon an unobservable physical system,
it must resolve a fundamental operational dilemma: which latent distinctions
actually govern the downstream task, how many active interventions are necessary
to certify them, and when must the system abstain rather than risk catastrophic
misclassification? In classical control and physical estimation, observability
is posed as an unconditioned binary predicate: either the microscopic state can
be uniquely reconstructed, or it is unobservable. In deployed environments,
however, passive observations cannot break latent degeneracies without active
perturbation, full microscopic inversion is prohibitively expensive, and
attempting to distinguish task-irrelevant degrees of freedom squanders bounded
interaction budgets. We formulate \emph{task-conditioned active observability
complexity}, determining the minimum worst-case expected interaction cost
required to identify the task-relevant current state under explicit error and
safe abstention guarantees. We prove that task-predictive equivalence induces the
unique minimal sufficient quotient $\mathcal{H}/\!\sim_\tau$, leaving active
observability complexity strictly invariant while eliminating superfluous
physical distinctions. In deterministic regimes, this complexity is
characterized exactly by an optimal adaptive distinguishing tree and a Bellman
recursion; in noisy regimes, it obeys a stopped-transcript relative-entropy
lower bound alongside adaptive martingale certificates that compose without
independence assumptions. We instantiate the theory in a prospective certified
observer that operationalizes staged recovery: a nominal verifier defers
candidate compilation, triggering active probing only upon evidence, while a
history-measurable score shell prunes hypotheses without sacrificing risk bounds.
Stress audits across high-dimensional physical systems and thousands of
operational trials demonstrate that the observer reliably recovers states while
slashing sensor reads and model steps, achieving zero false acceptances and safely
abstaining under ambiguity. The framework bridges discovered physical
ontologies and certified real-time control, transforming active state recovery
from unconditioned microscopic inversion into provably minimal physical
inquiry.
\end{abstract}

\section{Introduction}

In classical control and physical estimation, observability is posed as an
all-or-nothing predicate: either the microscopic state can be uniquely
reconstructed, or it is unobservable. In practical deployment, however, this
classical formulation overlooks critical operational constraints. Passive
observations often cannot break latent degeneracies without physical
perturbation; fully inverting microscopic state spaces is computationally and
energetically prohibitive; and downstream tasks typically require distinguishing
task-relevant operational regimes rather than resolving superfluous physical
degrees of freedom. Conversely, an observer that prematurely discards candidate
hypotheses risks catastrophic misclassification. Bridging this gap demands an
operational theory of observability that explicitly balances active interaction
costs, hypothesis memory, and certified decision risk.

We address this challenge by formalizing \emph{task-conditioned active observability}:
given a physical system with candidate hypotheses, how much active physical
interaction is necessary and sufficient to identify the task-relevant state
under certified error and safe abstention guarantees? In this setting, an
autonomous observer adaptively selects probe actions, updates belief over
candidate hypotheses, and terminates either with a certified task decision or
with a provably safe abstention when ambiguity cannot be resolved within budget.

Our first contribution formalizes task-conditioned active observability
complexity, separating representation complexity (which candidate distinctions
govern downstream task outcomes) from interaction complexity (the minimal cost
of adaptive experiments required to separate them). We prove that
task-predictive equivalence induces the unique minimal sufficient quotient
$\mathcal{H}/\!\sim_\tau$, eliminating task-irrelevant distinctions while leaving
active observability complexity strictly invariant. In deterministic settings,
this complexity is characterized exactly by an optimal adaptive distinguishing
tree via a Bellman recursion; in stochastic regimes, it satisfies a
stopped-transcript relative-entropy lower bound alongside adaptive martingale
certificates that compose constructively under arbitrary filtration histories.

Our second contribution is an operational observer architecture that realizes
this theoretical separation in practice. The observer decouples tracking from
recovery by verifying a lightweight nominal state first, compiling the broader
fallback bank only when accumulated evidence warrants, and certifying state
transitions on independent forward measurements. To scale recovery, a
history-measurable score-shell rule compresses the candidate bank while provably
preserving conditional safety guarantees, ensuring unresolvable ambiguities map
to certified abstention rather than misclassification. Furthermore, conformal
calibration of the shell rank controls marginal coverage, while an admissible
lower-bound certificate enables direct, closed-form candidate compilation.

Our third contribution establishes an optimality principle for physical resource
scheduling. We prove that when candidate compilation produces no useful
intermediate outputs during sensing, any interleaved scheduling between sensor
reads and partial compilation is pathwise weakly dominated by an atomic fallback
policy. This result establishes that compilation progress does not act as an
intrinsic state variable of the active observability problem, providing a formal
justification for executing fallback compilation atomically upon certified trigger.

We validate the theoretical framework across high-dimensional physical systems
and extensive operational trials. The experiments demonstrate that the staged
observer reliably recovers task-relevant states while substantially reducing
sensor queries and model computation, achieving zero false acceptances and
safely abstaining under ambiguity. Together, these results provide an end-to-end
framework for certified active state recovery, transforming classical
observability from unconditioned microscopic inversion into provably minimal
physical inquiry.

\section{Problem formulation}
\label{sec:problem}

\subsection{Finite controlled hypothesis model}

A completed operation history produces a finite candidate set
$\Hset=\Hset(H_0)$. A hypothesis $h\in\Hset$ contains a candidate operation
endpoint and its known future controlled transition and observation model. At
future time $t$, the observer selects $A_t\in\Aset$, receives
$Y_t\in\Yset$, and updates the propagated state of each candidate. The action
rule, stopping time $N$, and output are nonanticipating. Policies are
deterministic measurable maps of the observed history; there is no exogenous
randomization, and all probability comes from the observation process. The output
is either a label $\widehat\tau\in\Tset$ or abstention $\bot$.

A fixed task map
\begin{equation}
  \tau:\Hset\rightarrow\Tset
\end{equation}
specifies which distinctions matter. It can request the exact operation
endpoint, an active/inactive label, or any other fixed task-relevant quotient.
When the desired report is a current state, the operation-end label can be
propagated through the executed confirmation actions before output.

Let $c(a)>0$ be the physical interaction cost of action $a$. A policy $\pi$ is
$(\alpha,\beta)$-valid if, uniformly for all $h\in\Hset$,
\begin{align}
 \Pr_h^\pi\!\left(\widehat\tau\notin\{\tau(h),\bot\}\right)&\le \alpha,
 \label{eq:error}\\
 \Pr_h^\pi\!\left(\widehat\tau=\bot\right)&\le \beta.
 \label{eq:abstain}
\end{align}
The worst-case task-conditioned active observability complexity is
\begin{equation}
\AO_{\alpha,\beta}(\Hset,\tau)
=
\inf_{\pi\ \mathrm{valid}}
\sup_{h\in\Hset}
\mathbb E_h^\pi\!\left[\sum_{t=1}^{N}c(A_t)\right].
\label{eq:ao}
\end{equation}
The value is $+\infty$ if no valid policy exists. Computation is recorded
separately in the empirical study; it can also be folded into a generalized cost
when it is serial and action-independent.

\subsection{Task-predictive equivalence}

Let $\mathfrak E$ be the family of finite deterministic nonanticipating experimental policies.
For $\nu\in\mathfrak E$, write $P_h^\nu$ for the induced finite transcript law.

\begin{definition}[Task-predictive equivalence]
Two hypotheses are equivalent, written $h\equiv_\tau h'$, when
\begin{equation}
\tau(h)=\tau(h')
\quad\text{and}\quad
P_h^\nu=P_{h'}^\nu\quad\text{for every }\nu\in\mathfrak E.
\end{equation}
\end{definition}
For standard controlled Markov kernels, equality can equivalently be checked on
all finite open-loop action words; adaptive transcript equality then follows by
conditioning on the common history and induction.

\begin{theorem}[Minimal predictive quotient]
\label{thm:quotient}
Let $Q=\Hset/\!\equiv_\tau$. Then:
\begin{enumerate}
  \item every representation from which both $\tau(h)$ and all future
  experimental laws can be recovered must refine $Q$;
  \item $Q$ is sufficient;
  \item for every $\alpha,\beta$,
  \begin{equation}
  \AO_{\alpha,\beta}(\Hset,\tau)
  =\AO_{\alpha,\beta}(Q,\bar\tau).
  \end{equation}
\end{enumerate}
Thus $Q$ is the unique minimal sufficient representation up to relabeling.
\end{theorem}

The theorem isolates the only candidate distinctions that can affect either the
answer or any future evidence. It also justifies quotienting before optimizing
an active observer.

\begin{corollary}[Identifiability obstruction]
\label{cor:obstruction}
Suppose two hypotheses have different task labels but identical transcript laws
under every policy. If $2\alpha+\beta<1$, then no
$(\alpha,\beta)$-valid observer exists.
\end{corollary}

\section{Active observability theory}
\label{sec:theory}

\subsection{Exact deterministic characterization}

Assume deterministic transitions and observations. An information configuration
$S$ is a finite set of pairs $(\ell,z)$, where $\ell$ is the fixed task label of
an initial hypothesis and $z$ is its current propagated state. Duplicate pairs
are removed. A configuration is \emph{pure} if all pairs have the same label.
For action $a$ and observation $y$, let $S_{a,y}$ be the propagated subset that
produces $y$.

A task-separating experiment tree places actions at internal nodes, observations
on edges, and label-pure configurations at leaves. Define its minimax cost by
\begin{equation}
\DT(S)=
\inf_{\mathcal T}
\max_{\ell\in\operatorname{leaves}(\mathcal T)}
\sum_{v\in\operatorname{path}(\ell)}c(a_v),
\label{eq:dtree}
\end{equation}
with $\DT(S)=+\infty$ when no finite separating tree exists.

\begin{theorem}[Deterministic closure]
\label{thm:deterministic}
For a deterministic noiseless controlled system,
\begin{equation}
\AO_{0,0}(S)=\DT(S).
\end{equation}
Equivalently, $\DT$ is the least extended nonnegative solution of
\begin{equation}
\DT(S)=0\quad\text{for pure }S,
\end{equation}
and otherwise
\begin{equation}
\DT(S)=
\inf_{a\in\Aset}
\left\{c(a)+
\max_{y:S_{a,y}\ne\varnothing}\DT(S_{a,y})
\right\}. 
\label{eq:bellman}
\end{equation}
\end{theorem}

The equality is exact, not asymptotic: every exact policy unfolds into a
separating tree, and every such tree is an executable exact policy. Combined
with \cref{thm:quotient}, it gives a two-step characterization: first quotient
the representation, then solve the minimum distinguishing-tree problem.

\subsection{Noisy information lower bound}

Let $\mathbb P_h^\pi$ denote the stopped transcript law, including the decision,
under policy $\pi$. Define binary relative entropy
\begin{equation}
 d(p\|q)=p\log\frac pq+(1-p)\log\frac{1-p}{1-q}.
\end{equation}

\begin{theorem}[Pairwise stopped-transcript necessity]
\label{thm:kl}
Assume $2\alpha+\beta<1$. For every $(\alpha,\beta)$-valid policy and every
pair with $\tau(h)\ne\tau(h')$,
\begin{equation}
\KL\!\left(\mathbb P_h^\pi\middle\|\mathbb P_{h'}^\pi\right)
\ge d(1-\alpha-\beta\|\alpha).
\label{eq:kl-event}
\end{equation}
\end{theorem}

For controlled Markov observations with kernel $K(\cdot\mid z,a)$, the adaptive
chain rule yields
\begin{equation}
\begin{aligned}
\KL\!\left(\mathbb P_h^\pi\middle\|\mathbb P_{h'}^\pi\right)
&=\mathbb E_h^\pi\sum_{t=1}^{N}
\KL\!\left(K(\cdot\mid z_t^h,A_t)
\middle\|\right.\\[-0.2em]
&\hspace{7.3em}\left.K(\cdot\mid z_t^{h'},A_t)\right).
\end{aligned}
\label{eq:chain}
\end{equation}
Let $\Gamma_{h\to h'}$ be the largest one-step directed KL per unit cost over
paired states reachable under a common action history. Then
\begin{equation}
\begin{aligned}
\AO_{\alpha,\beta}(\Hset,\tau)
&\ge \max_{\tau(h)\ne\tau(h')}
\max\!\left\{
\frac{d(1-\alpha-\beta\|\alpha)}{\Gamma_{h\to h'}},\right.\\[-0.2em]
&\hspace{7.1em}\left.
\frac{d(1-\alpha-\beta\|\alpha)}{\Gamma_{h'\to h}}
\right\}.
\end{aligned}
\label{eq:kl-lower}
\end{equation}
A zero information rate for any task-inequivalent pair makes the complexity
infinite.

\subsection{Certified-tree upper bound}

A node-level certified test may itself be sequential. At node $v$ with candidate
configuration $S_v$, it has a correct branch map $b_v:S_v\to B_v$ and, conditional
on every possible entry history, obeys
\begin{align}
\Pr_h(\widehat b_v\notin\{b_v(h),\bot\}\mid\Fhist_{\mathrm{entry}})&\le\alpha_v,\\
\Pr_h(\widehat b_v=\bot\mid\Fhist_{\mathrm{entry}})&\le\beta_v,
\end{align}
with conditional expected cost at most $c_v$. A certified distinguishing tree
has task-pure leaves and stops globally if any node abstains.

\begin{theorem}[Adaptive certified composition]
\label{thm:composition}
Executing a certified distinguishing tree gives
\begin{align}
\Pr(\text{wrong task label})
&\le
\max_{\ell}\sum_{v\in\operatorname{path}(\ell)}\alpha_v,\\
\Pr(\text{abstain})
&\le
\max_{\ell}\sum_{v\in\operatorname{path}(\ell)}\beta_v,
\end{align}
and worst-case expected cost at most
\begin{equation}
\max_{\ell}\sum_{v\in\operatorname{path}(\ell)}c_v.
\end{equation}
No independence between node tests is required.
\end{theorem}

\begin{corollary}[Active-observability sandwich]
\label{cor:sandwich}
Let $\mathsf L_{\alpha,\beta}$ denote the pairwise information lower bound in
\cref{eq:kl-lower}, and let $\mathsf U_{\alpha,\beta}$ be the least path cost of
a certified tree satisfying the pathwise risk budgets. Then
\begin{equation}
\mathsf L_{\alpha,\beta}
\le \AO_{\alpha,\beta}
\le \mathsf U_{\alpha,\beta}.
\end{equation}
In the deterministic noiseless specialization, the certified-tree optimum reduces to the exact quantity $\\DT$ in \\cref{thm:deterministic}.
\end{corollary}

\section{Certified observer construction}
\label{sec:observer}

\subsection{Nominal-first staged confirmation}

Rather than compiling the full hypothesis space $\mathcal{H}(H_0)$ upfront, the observer maintains a nominal candidate $N=\{n\}$ for expected tracking while deferring compilation of the generic fallback candidate bank $G(H_0)$, where the unified candidate set is $U=N\cup G$. The observer first interrogates $N$ on a fresh observation stream under risk budget $\alpha_{\mathrm{nom}}$. If confirmed, $n$ is accepted immediately at minimal query and compute cost; otherwise, atomic fallback compiles and verifies $U$ on subsequent independent measurements under risk budget $\alpha_{\mathrm{fall}}$. Partitioning the overall risk budget as $\alpha_{\mathrm{nom}}+\alpha_{\mathrm{fall}}\le\alpha$ guarantees conditional safety via \cref{thm:composition}, with candidate set instantiation statistically decoupled from certification.

\subsection{Readwise evidence-triggered fallback}

In latency-sensitive deployments, waiting for exhaustive nominal rejection can incur unnecessary physical query cost. The observer incorporates an evidence-triggered sequential rule that monitors the running verification transcript readwise. When intermediate observations yield strong discordant evidence against $N$, the system preemptively triggers atomic fallback without exhausting the nominal testing budget. Because the trigger functions as an adaptive stopping rule while fallback certification operates on disjoint subsequent data, this early switching modulates interaction cost without compromising conditional safety guarantees. The parameter $\lambda \ge 0$ governs the interaction-versus-compilation tradeoff in \cref{alg:observer}, evaluated across frozen values $\lambda\in\{0,10,100,1000\}$ alongside eager and staged baselines.

\begin{algorithm}[t]
\caption{Certified Staged Active Observer}
\label{alg:observer}
\begin{algorithmic}[1]
\REQUIRE Tracking history $H_0$, task map $\tau$, query budget $T_{\max}$, threshold $\lambda$, risk budgets $(\alpha_{\mathrm{nom}}, \alpha_{\mathrm{fall}})$.
\ENSURE Certified task decision $\hat{\ell} \in \Theta$ or safe abstention $\bot$.
\STATE Initialize nominal candidate $N \leftarrow \{n\}$ from history $H_0$.
\STATE \textbf{// Phase 1: Nominal sequential verification}
\FOR{$t = 1, \dots, T_{\max}$}
    \STATE Query action $A_t$ and record measurement $Y_t$.
    \STATE Update nominal evidence margin $\Delta_t(n)$.
    \IF{$\Delta_t(n) \ge \gamma_{\mathrm{conf}}(\alpha_{\mathrm{nom}})$}
        \RETURN $\tau(n)$ \COMMENT{Nominal certified under risk $\alpha_{\mathrm{nom}}$}
    \ELSIF{$\Delta_t(n) \le -\lambda$}
        \STATE \textbf{break} \COMMENT{Preemptively trigger fallback}
    \ENDIF
\ENDFOR
\STATE \textbf{// Phase 2: Score-shell candidate compression}
\STATE Instantiate generic bank $G(H_0)$ and form union $U \leftarrow N \cup G(H_0)$.
\STATE Evaluate lexicographic scores $\rho(c)$ via \eqref{eq:score} for $c \in U$.
\STATE Compress to top-$J$ score-shells: $S_{\mathrm{shell}} \leftarrow \operatorname{FilterShells}(U, J=8)$.
\STATE \textbf{// Phase 3: Disjoint confirmation and certified decision}
\STATE Execute verification actions on $S_{\mathrm{shell}}$ over fresh observation stream.
\IF{unique candidate $c^* \in S_{\mathrm{shell}}$ is certified under risk $\alpha_{\mathrm{fall}}$}
    \RETURN $\tau(c^*)$
\ELSE
    \RETURN $\bot$ \COMMENT{Safe abstention satisfying \cref{thm:composition}}
\ENDIF
\end{algorithmic}
\end{algorithm}

\subsection{Score-shell compression and exact compilation}
\label{subsec:score-shell}

Each endpoint candidate $c$ receives the lexicographic score
\begin{equation}
\rho(c)=
\left(
\operatorname{mis}(c),
\operatorname{so}(c)+\operatorname{sy}(c),
\max\{\operatorname{so}(c),\operatorname{sy}(c)\}
\right).
\label{eq:score}
\end{equation}
Candidates with the same score form a shell. The compressed representation keeps
the first $J$ distinct shells after endpoint-wise minimization, plus the nominal
endpoint.

\begin{theorem}[History-measurable sub-bank safety]
\label{thm:subbank}
Let $H_0$ be the completed operation history and let $B(H_0)$ be any finite
history-measurable candidate bank fixed before a fresh confirmation stream. If
the verifier has conditional family-wise false-accept probability at most
$\alpha$ for every fixed bank, then
\begin{equation}
\Pr(\text{wrong acceptance}\mid H_0)\le\alpha.
\end{equation}
Candidate omission can reduce coverage, but it cannot increase the allocated
false-accept bound.
\end{theorem}

Calibration is performed in exchangeable blocks that each contain all 20 active
strata. If $M_i$ is the maximum true-shell rank in block $i$ and
$J_m=\max_{i\le m}M_i$, then
\begin{equation}
\Pr(M_{m+1}\le J_m)\ge\frac{m}{m+1}.
\label{eq:block}
\end{equation}
The inequality follows because failure requires the final block to be a unique
strict maximum. With $m=100$, the frozen choice $J=8$ has next-block marginal
coverage at least $100/101$.

\paragraph{Exact direct compilation.}
To avoid enumerating the full bank before discarding later shells, every partial search node $u$ is assigned an admissible lexicographic lower bound $L(u)$ on the score of all of its completions. Once eight distinct endpoint-best shells have been found with eighth threshold $\theta_8$, the search terminates directly when
\begin{equation}
\min_{u\in\mathrm{OPEN}}L(u)>_{\mathrm{lex}}\theta_8.
\label{eq:direct-cert}
\end{equation}
When this certificate fails, the compiler falls back to exact exhaustive enumeration. Hence the hybrid compiler returns the identical endpoint-score candidate map as exhaustive filtering on every input while bypassing unviable branches.

\begin{theorem}[Atomic fallback dominance]
\label{thm:atomic}
Assume the compilation target is fixed before confirmation, compilation has no
usable intermediate output, compute produces no plant observation, read and
compute costs add serially, and there is no deadline, discount, or reward for
early completion. Every policy that interleaves reads and partial compilation is
pathwise weakly dominated by a policy that performs no compilation until an
observation stopping time and then either never compiles or completes the entire
fallback atomically.
\end{theorem}

The proof moves all useful compile microsteps to the first point at which the completed bank is used and deletes all unused work. The read sequence, decision, and useful computation are unchanged; therefore, compilation progress need not appear as an additional state variable in the active-observability formulation.

\section{Empirical evaluation}
\label{sec:experiments}

\subsection{Experimental setup and evaluation metrics}
\label{subsec:protocol}

We evaluate the certified active observability framework across three complementary benchmark domains:

\paragraph{Benchmark domains.}
First, a prospective closed-loop control testbed evaluates sequential tracking and fallback policies across two 24-bit controlled Boolean models with 96 public commands. Each operational episode executes 192 commanded actions under hidden command substitutions and sensor flips. The benchmark spans 72 prospective tasks across six families, consisting of 20 active and 52 inactive operations. To ensure strict paired comparisons without multiplying physical risk, every policy replays identical prospective observation streams of length at most 600.
Second, a large-scale confirmatory benchmark evaluates candidate compression and compiler exactness over 100 calibration blocks and 100 independent evaluation blocks, comprising 2,000 total operations across the 20 active strata. All hyperparameters are frozen prior to evaluation, with $J=8$, a probe target of 16 shells, and an evaluation budget of 95,000 queries.
Third, a synthetic control suite evaluates theoretical invariants across 600 random deterministic Mealy automata and 5,000 adaptive stochastic policies.

\paragraph{Evaluation dimensions.}
Performance is audited across four operational axes:
(i)~\textbf{physical interaction cost}, measured by future sensor reads and commanded actions;
(ii)~\textbf{computational complexity}, quantified by candidate-compilation evaluations and verifier model steps;
(iii)~\textbf{certification reliability}, tracked via correct confirmations, safe abstentions, and empirical false acceptances; and
(iv)~\textbf{compiler exactness}, verified by state-by-state equality of pruned versus full candidate maps.
Candidate coverage and conditional false-accept safety are audited and reported separately.

\subsection{Closed-loop interaction and computation trade-offs}
\label{subsec:controller}

\begin{figure}[t]
\centering
\includegraphics[width=0.82\linewidth]{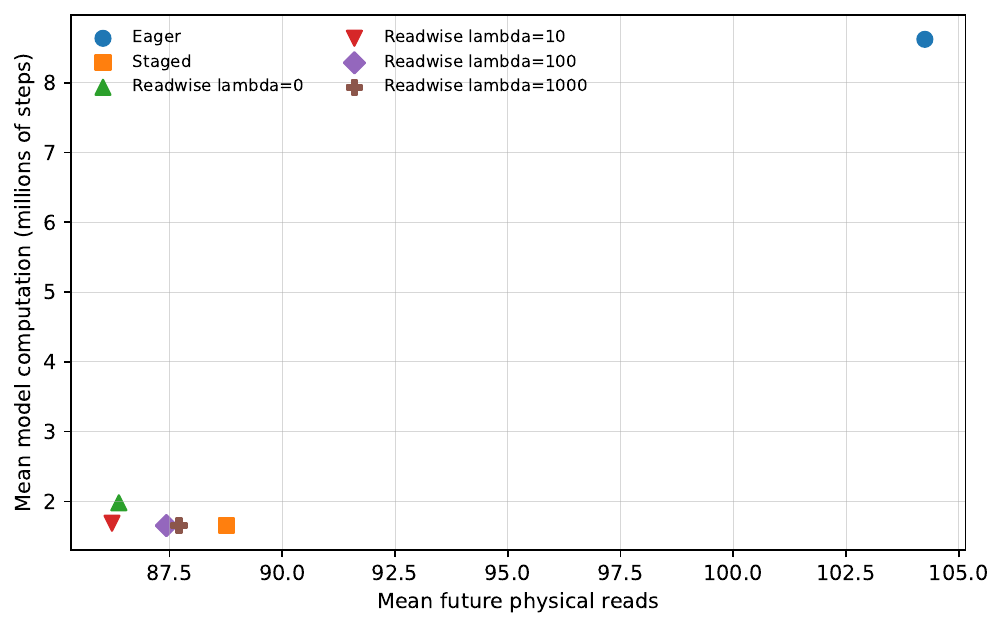}
\caption{Mean physical interaction versus mean model computation for the six
frozen controller policies. Eager compilation is separated from the cluster of
deferred policies. Readwise thresholds trace a physical-read/computation
tradeoff inside that cluster.}
\label{fig:controller}
\end{figure}

All six evaluated controller policies correctly confirm all 72 prospective tasks without producing unknown outcomes. \Cref{tab:controller} and \Cref{fig:controller} summarize the aggregate interaction and computational trade-offs. Eager compilation incurs severe computational waste by executing 8.62 million model steps to compile candidates unconditionally across all 52 inactive tasks. In contrast, nominal-first staging exploits expected physical regularity, reducing model computation more than five-fold to 1.65 million steps while lowering mean future reads from 104.25 to 88.76.

\begin{table*}[t]
\centering
\caption{Prospective task-level controller results. All policies are correct on
72/72 tasks. Model computation includes candidate compilation, propagation, and
verifier steps.}
\label{tab:controller}
\begin{tabular}{lrrrr}
\toprule
Policy & Mean reads & Model steps (M) & Compilations & Correct \\
\midrule
Eager & 104.250 & 8.6213 & 72 & 72/72 \\
Staged & 88.764 & 1.6532 & 20 & 72/72 \\
Readwise $\lambda=0$ & 86.375 & 1.9802 & 47 & 72/72 \\
Readwise $\lambda=10$ & \textbf{86.222} & 1.6862 & 36 & 72/72 \\
Readwise $\lambda=100$ & 87.431 & 1.6527 & 21 & 72/72 \\
Readwise $\lambda=1000$ & 87.708 & \textbf{1.6505} & \textbf{20} & 72/72 \\
\bottomrule
\end{tabular}
\end{table*}

Readwise evidence-triggered switching further optimizes this Pareto frontier. Rather than waiting to exhaust the nominal testing budget, the readwise rule monitors accumulated discordance and triggers fallback early. Conservative switching with $\lambda=1000$ compiles the exact same 20 active tasks as formal staging while trimming mean future reads from 88.76 to 87.71. More aggressive switching with $\lambda=10$ achieves the lowest physical interaction at 86.22 reads, though incurring 16 unnecessary compilations on inactive tasks.

\begin{table}[t]
\centering
\footnotesize
\caption{Stratified performance breakdown across the 72 prospective controller tasks under formal staging and conservative readwise fallback.}
\label{tab:strata}
\setlength{\tabcolsep}{2pt}
\begin{tabular}{@{}l>{\raggedright\arraybackslash}p{0.26\columnwidth}rrr@{}}
\toprule
Outcome & Policy & Episodes & Mean reads & Compilations \\
\midrule
Active & Staged & 20 & 104.600 & 20 \\
Active & Readwise $\lambda=1000$ & 20 & 100.800 & 20 \\
Inactive & Staged & 52 & 82.673 & 0 \\
Inactive & Readwise $\lambda=1000$ & 52 & 82.673 & 0 \\
\bottomrule
\end{tabular}
\end{table}

\Cref{tab:strata} reveals the mechanistic origin of these savings by stratifying tasks across operational outcomes. The domain naturally bifurcates into two regimes:
(i)~\textbf{inactive operations}, comprising 52 tasks or 72.2\% of the domain, where the nominal hypothesis is confirmed immediately using only 82.67 future reads and zero candidate compilations; and
(ii)~\textbf{active operations}, comprising 20 tasks or 27.8\% of the domain, where conservative readwise triggering with $\lambda=1000$ reduces average reads from 104.60 to 100.80. Crucially, the median fallback trigger shifts from 9 reads under formal rejection down to 4 reads under readwise monitoring. Evidence-triggered switching thus protects nominal-path throughput without compromising worst-case safety auditing.

\subsection{Representation compression and exact compilation}
\label{subsec:compression}

\begin{figure}[t]
\centering
\includegraphics[width=0.84\linewidth]{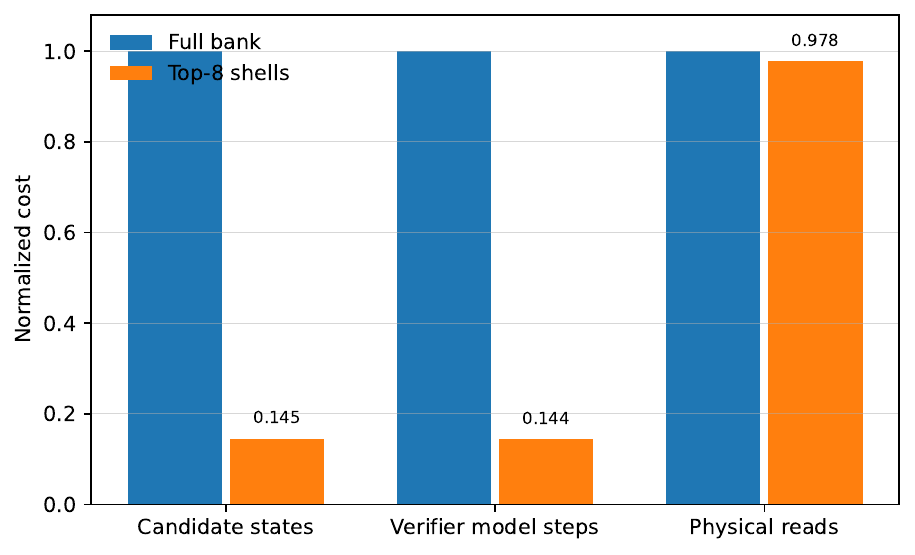}
\caption{Cost retained after replacing the full fallback bank with the first
eight score shells. Candidate states and verifier computation fall to about
14.5\% of the full-bank values; physical interaction changes much less.}
\label{fig:shell}
\end{figure}

\paragraph{Candidate compression and coverage calibration.}
Restricting the fallback bank to the top-$J=8$ score shells eliminates 85.51\% of candidate states, dropping from 7,520,508 to 1,090,024, and 85.57\% of verifier model steps, dropping from 140,052,481 to 20,210,639, while reducing physical reads by 2.19\% through earlier verification (\Cref{tab:shell}, \Cref{fig:shell}).
Across 2,000 independent operations, the true endpoint is certified in 1,999 instances, with the single omission resulting in a safe abstention, thereby achieving zero false acceptances and zero unknown outcomes. This confirms \cref{thm:subbank}: representation pruning governs empirical coverage as a calibrated availability property while strictly preserving conditional safety.

\begin{table*}[t]
\centering
\caption{Independent 2,000-operation score-shell confirmation.}
\label{tab:shell}
\begin{tabular}{lrrr}
\toprule
Metric & Full bank & Top-8 shells & Change \\
\midrule
Candidate states & 7,520,508 & 1,090,024 & $-85.51\%$ \\
Verifier model steps & 140,052,481 & 20,210,639 & $-85.57\%$ \\
Physical reads & 203,135 & 198,683 & $-2.19\%$ \\
Correct confirmations & 2,000 & 1,999 & $-1$ \\
Safe rejections & 0 & 1 & $+1$ \\
Wrong acceptances & 0 & 0 & 0 \\
\bottomrule
\end{tabular}
\end{table*}

\paragraph{Compiler exactness and search truncation.}
Direct score-shell compilation reproduces the candidate-score map of full exhaustive enumeration with 100\% state-by-state identity across all 2,000 operations.
When the lexicographic bound in \eqref{eq:direct-cert} certifies that unexpanded branches cannot reach the eighth score shell, search terminates early. Direct certificates are established in 38 operations, reducing compilation evaluations by 28.19\% on those instances. Over the full 2,000-operation benchmark, total evaluations fall from 10,033,791,134 to 9,929,092,066, representing a net reduction of 104,699,068 evaluations (1.04\%) that fully absorbs the 91,390,153 evaluations expended on uncertified exploratory probes in the remaining 962 operations. Exact fallback boundaries ensure that branch pruning introduces zero approximation error into state verification.

\paragraph{Theoretical invariants on synthetic systems.}
Across 600 random Mealy automata and 5,000 adaptive noisy policies, the structural predictions of \cref{sec:theory} hold uniformly. Task-predictive quotienting preserves exact distinguishing depth in all 600 deterministic systems while reducing state representations in 75. In stochastic regimes, stopped-transcript relative entropy strictly dominates the divergence of every induced binary decision event, and the explicit information lower bound $d(1-\alpha-\beta\|\alpha)$ holds across all 461 nonvacuous risk-constrained instances.

\section{Related work}
\label{sec:related}

Classical linear and nonlinear observability characterize whether internal state
can be reconstructed from input-output behavior
\citep{kalman1960new,hermann1977nonlinear}. Active state-estimation work chooses
inputs to improve estimation quality \citep{hu2004active}. Our formulation differs
by making task equivalence, finite candidate representations, interaction cost,
error, and abstention explicit, and by seeking an exact finite-system
characterization rather than a rank condition or estimator design.

Sequential experimental design and controlled sensing study how actions should be
chosen to discriminate hypotheses under sample or decision-risk constraints
\citep{chernoff1959sequential,wald1947sequential,naghshvar2013active,nitinawarat2013controlled}.
The KL lower bound in \cref{thm:kl} follows this information-theoretic lineage.
Our emphasis is the interface between evolving controlled states, task-conditioned
quotients, certified node tests, candidate compilation, and safe abstention.

Finite-state-machine testing studies state identification, distinguishing
sequences, and conformance experiments
\citep{moore1956gedanken,chow1978testing,lee1994testing,vandenbos2019state}.
The deterministic tree in \cref{thm:deterministic} is closely related to adaptive
distinguishing experiments, but here it is embedded in a risk-controlled noisy
observer and coupled to history-dependent candidate representations.

While automata learning reconstructs unknown state machines from query responses \citep{angluin1987learning,vaandrager2017model}, our setting optimizes state recovery within a known dynamical representation.

\paragraph{Interface with physical coordinate discovery and ontology learning.}

The active observability framework developed here addresses a complementary operational challenge to the physical representation learning program established in Papers~1--3 of this series \citep{zhang2026representation,zhang2026thermodynamic,zhang2026threefreq}.  In physical representation learning, an autonomous agent seeks to discover the continuous coordinate geometry, Lie group gauge symmetries, and thermodynamic contact structures $(\mathcal{T}, \eta)$ of an unknown physical system through experimental scaling, contact, and reservoir interventions. Once this continuous physical ontology is discovered and frozen, practical execution in autonomous facilities, such as self-driving synthesis laboratories or battery management platforms, requires monitoring the operational regime in real time.

In such deployed environments, full continuous state deconvolution is often unobservable or prohibitively expensive under bounded instrumentation.  Task-conditioned active observability bridges this divide: by quotienting the state space modulo the target control task into $\mathcal{H} / \!\sim_\tau$, the agent avoids wasting physical resources to resolve task-irrelevant physical fluctuations, while retaining certified mathematical guarantees against erroneous state assertions.

\section{Limitations}
\label{sec:limits}

The theoretical guarantees apply to finite controlled models with known representations under deterministic policies. The formulation focuses on certified tracking within verified dynamics and does not address unanchored cold-start identification or unmodeled process faults. Extending the guarantees from marginal next-block coverage to distribution-free conditional coverage across arbitrary covariates remains an open theoretical direction.

Our atomicity analysis considers sequential evaluation; extending the framework to parallel execution, execution deadlines, or incremental candidate generation presents a practical direction for deployment.

\section{Conclusion}

{\looseness=-1
Task-conditioned active observability separates a system's minimal predictive
representation from the interaction needed to identify its current task-relevant
state. The predictive quotient is uniquely minimal and leaves the complexity
unchanged. Deterministic systems admit an exact adaptive-tree characterization;
noisy systems admit complementary information lower bounds and certified-tree
upper bounds. The prospective observer demonstrates how these ideas organize
practical choices: defer expensive candidates, trigger fallback from evidence,
use fresh confirmation data after selection, compress representations without
conflating safety and coverage, and stop expanding the controller when partial
computation has no decision value.\par}

Within a fixed representation, tightening the gap between the noisy information lower bound and executable certified trees remains the primary open direction.

\FloatBarrier\clearpage
\bibliographystyle{icml2026}
\bibliography{references}

@article{kalman1960new,
  author  = {Kalman, Rudolf E.},
  title   = {A New Approach to Linear Filtering and Prediction Problems},
  journal = {Journal of Basic Engineering},
  volume  = {82},
  number  = {1},
  pages   = {35--45},
  year    = {1960},
  doi     = {10.1115/1.3662552}
}

@article{hermann1977nonlinear,
  author  = {Hermann, Robert and Krener, Arthur J.},
  title   = {Nonlinear Controllability and Observability},
  journal = {IEEE Transactions on Automatic Control},
  volume  = {22},
  number  = {5},
  pages   = {728--740},
  year    = {1977},
  doi     = {10.1109/TAC.1977.1101601}
}

@article{hu2004active,
  author  = {Hu, Xiaoming and Ersson, Thomas},
  title   = {Active State Estimation of Nonlinear Systems},
  journal = {Automatica},
  volume  = {40},
  number  = {12},
  pages   = {2075--2082},
  year    = {2004},
  doi     = {10.1016/j.automatica.2004.06.015}
}

@article{chernoff1959sequential,
  author  = {Chernoff, Herman},
  title   = {Sequential Design of Experiments},
  journal = {The Annals of Mathematical Statistics},
  volume  = {30},
  number  = {3},
  pages   = {755--770},
  year    = {1959},
  doi     = {10.1214/aoms/1177706205}
}

@article{naghshvar2013active,
  author  = {Naghshvar, Mohammad and Javidi, Tara},
  title   = {Active Sequential Hypothesis Testing},
  journal = {The Annals of Statistics},
  volume  = {41},
  number  = {6},
  pages   = {2703--2738},
  year    = {2013},
  doi     = {10.1214/13-AOS1144}
}

@article{nitinawarat2013controlled,
  author  = {Nitinawarat, Sirin and Atia, George K. and Veeravalli, Venugopal V.},
  title   = {Controlled Sensing for Multihypothesis Testing},
  journal = {IEEE Transactions on Automatic Control},
  volume  = {58},
  number  = {10},
  pages   = {2451--2464},
  year    = {2013},
  doi     = {10.1109/TAC.2013.2261188}
}

@book{wald1947sequential,
  author    = {Wald, Abraham},
  title     = {Sequential Analysis},
  publisher = {John Wiley and Sons},
  address   = {New York},
  year      = {1947}
}

@incollection{moore1956gedanken,
  author    = {Moore, Edward F.},
  title     = {Gedanken-Experiments on Sequential Machines},
  booktitle = {Automata Studies},
  editor    = {Shannon, Claude E. and McCarthy, John},
  series    = {Annals of Mathematics Studies},
  volume    = {34},
  pages     = {129--153},
  publisher = {Princeton University Press},
  year      = {1956}
}

@article{chow1978testing,
  author  = {Chow, Tsun S.},
  title   = {Testing Software Design Modeled by Finite-State Machines},
  journal = {IEEE Transactions on Software Engineering},
  volume  = {SE-4},
  number  = {3},
  pages   = {178--187},
  year    = {1978},
  doi     = {10.1109/TSE.1978.231496}
}

@article{lee1994testing,
  author  = {Lee, David and Yannakakis, Mihalis},
  title   = {Testing Finite-State Machines: State Identification and Verification},
  journal = {IEEE Transactions on Computers},
  volume  = {43},
  number  = {3},
  pages   = {306--320},
  year    = {1994},
  doi     = {10.1109/12.272431}
}

@article{vandenbos2019state,
  author  = {van den Bos, Petra and Vaandrager, Frits},
  title   = {State Identification for Labeled Transition Systems with Inputs and Outputs},
  journal = {arXiv preprint arXiv:1907.11034},
  year    = {2019},
  eprint  = {1907.11034},
  archivePrefix = {arXiv},
  primaryClass  = {cs.LO}
}

@article{angluin1987learning,
  author  = {Angluin, Dana},
  title   = {Learning Regular Sets from Queries and Counterexamples},
  journal = {Information and Computation},
  volume  = {75},
  number  = {2},
  pages   = {87--106},
  year    = {1987},
  doi     = {10.1016/0890-5401(87)90052-6}
}

@article{vaandrager2017model,
  author  = {Vaandrager, Frits},
  title   = {Model Learning},
  journal = {Communications of the ACM},
  volume  = {60},
  number  = {2},
  pages   = {86--95},
  year    = {2017},
  doi     = {10.1145/2967606}
}

@article{zhang2026representation,
  author  = {Zhang, Linzhe and Xu, Changming},
  title   = {Discovering Physical Representation Languages: Metric-Free Axiomatics and Topological Ontology Identification},
  journal = {arXiv preprint},
  year    = {2026}
}

@article{zhang2026thermodynamic,
  author  = {Zhang, Linzhe and Xu, Changming},
  title   = {Blind Thermodynamic Ontology Discovery from Anonymous Experiments},
  journal = {arXiv preprint},
  year    = {2026}
}

@article{zhang2026threefreq,
  author  = {Zhang, Linzhe and Xu, Changming},
  title   = {Parasitic-Free Three-Frequency Laws for Positive Relaxation Ports},
  journal = {arXiv preprint},
  year    = {2026}
}

\FloatBarrier\clearpage
\appendix
\renewcommand{\thetable}{A\arabic{table}}
\renewcommand{\thefigure}{A\arabic{figure}}
\setcounter{table}{0}
\setcounter{figure}{0}

\section{Proofs}
\label{app:proofs}

\subsection{Proof of the minimal predictive quotient theorem}

Suppose a representation $r(h)$ is sufficient to recover both $\tau(h)$ and the
transcript law of every future experiment. If $r(h)=r(h')$, every decoded
quantity agrees, hence $\tau(h)=\tau(h')$ and $P_h^\nu=P_{h'}^\nu$ for every
$\nu$. Therefore $h\equiv_\tau h'$ and every sufficient representation refines
$Q$.

Conversely, one task label and one family of future experimental laws are
well-defined on every equivalence class, so $Q$ is sufficient. Any policy on
$\Hset$ induces identical transcript, cost, error, and abstention laws for all
members of a class and therefore descends to $Q$. Every policy on $Q$ lifts to
$\Hset$. Taking the same infimum and supremum proves complexity invariance.
\hfill$\square$

For \cref{cor:obstruction}, let $E$ be the event that the observer outputs
$\tau(h)$. Under $h$, validity requires
$\Pr_h(E)\ge1-\alpha-\beta$. Under $h'$, the same output is wrong, so
$\Pr_{h'}(E)\le\alpha$. Identical transcript laws imply identical output laws,
contradicting $1-\alpha-\beta>\alpha$.

\subsection{Proof of the deterministic closure theorem}

Fix any exact policy. By the policy convention in \cref{sec:problem}, its
decisions are deterministic functions of the observation history, so unfolding
them gives an action-observation tree. At every reachable stopping leaf, all
remaining candidate pairs must have the same task label; otherwise the common
output would be wrong for at least one candidate. Hence the policy induces a
finite task-separating tree with the same worst-case cost and cannot beat
$\DT(S)$.

Conversely, execute any task-separating tree and output the unique label at the
reached leaf. The policy is exact and has the tree's worst-case path cost. Taking
infima proves $\AO_{0,0}(S)=\DT(S)$. Decomposing a nonterminal tree at its root
gives the Bellman lower bound in \cref{eq:bellman}; adjoining optimal child trees
gives the reverse bound. Because all action costs are positive, the least
extended solution assigns $+\infty$ exactly to configurations with no finite
separating tree. \hfill$\square$

\subsection{Proof of the KL lower bound}

Let $E=\{\widehat\tau=\tau(h)\}$. Under $h$,
$\mathbb P_h^\pi(E)\ge1-\alpha-\beta$; under $h'$,
$\mathbb P_{h'}^\pi(E)\le\alpha$. Data processing through the indicator
$\ind_E$ gives
\begin{equation}
\KL(\mathbb P_h^\pi\|\mathbb P_{h'}^\pi)
\ge d(\mathbb P_h^\pi(E)\|\mathbb P_{h'}^\pi(E)).
\end{equation}
Since $1-\alpha-\beta>\alpha$, binary relative entropy is increasing in its first
argument and decreasing in its second over the relevant rectangle. This proves
\cref{eq:kl-event}.

For the cost bound, the action-selection kernels are the same functions of the
observed history under both hypotheses and cancel in the likelihood ratio. The
chain rule gives \cref{eq:chain}. Each term is at most
$\Gamma_{h\to h'}c(A_t)$, hence
\begin{equation}
\mathbb E_h^\pi\sum_{t=1}^N c(A_t)
\ge
\frac{d(1-\alpha-\beta\|\alpha)}{\Gamma_{h\to h'}}.
\end{equation}
Repeat in the reverse direction and use the worst-case objective.
\hfill$\square$

\subsection{Proof of certified adaptive composition}

A wrong terminal label implies that at least one visited node misrouted the true
hypothesis. Condition on each node's entry sigma-field. Its misrouting probability
is at most $\alpha_v$ for every possible entry history. The tower property and a
union bound over the realized root-to-leaf path give the path sum. Maximizing over
paths gives the first bound. The abstention proof is identical. Conditional
expected costs add by iterated expectation, and the sum along every realized path
is bounded by the largest path sum. No cross-node independence is used.
\hfill$\square$

\subsection{Proof of history-measurable sub-bank safety}

Condition on the completed operation history $H_0$. Then $B(H_0)$ is a fixed
finite bank. By the fixed-bank verifier guarantee,
\begin{equation}
\Pr(\text{wrong acceptance}\mid H_0,B(H_0))\le\alpha.
\end{equation}
Because $B(H_0)$ is measurable with respect to $H_0$, conditioning on both is the
same as conditioning on $H_0$. Candidate omission does not alter this inequality;
it only removes the completeness premise needed to guarantee an acceptance.
\hfill$\square$

For \cref{eq:block}, exchangeability makes every block equally likely to be the
unique strict maximum among $m+1$ block maxima. Failure occurs only when the last
block is that unique maximum, whose probability is at most $1/(m+1)$. Ties can
only improve coverage.

\subsection{Proof of atomic fallback dominance}

Fix a realized read/outcome path of an arbitrary policy. If no completed bank is
used, delete every compile microstep. If a completed bank is first used at some
observation stopping history, move every useful compile microstep to one
contiguous block immediately before that use. Compilation emits no observation,
does not change the plant, and has a fixed target, so all read-dependent
decisions and the completed bank remain unchanged. Additive serial cost remains
the same; unused work is deleted. Applying this exchange pathwise yields a policy
that never maintains partial progress and is weakly cheaper on every path.
\hfill$\square$

\section{Additional protocol details}
\label{app:protocol}

\subsection{Staged risk accounting}

The eager verifier uses false-accept budget $1/2{,}000{,}000$. The staged method
uses $\alpha_1=\alpha_2=1/4{,}000{,}000$. Let $A_1$ be a wrong nominal
acceptance and $A_2$ a wrong fallback acceptance. Conditional on the entire
stage-1 history and on entering stage 2, the propagated bank is fixed before the
fresh stage-2 stream. Therefore
\begin{equation}
\Pr(A_1\cup A_2)
\le\Pr(A_1)+\mathbb E[\Pr(A_2\mid\Fhist_{\mathrm{stage1}})]
\le\alpha_1+\alpha_2.
\end{equation}
This argument does not assume that the full or compressed bank contains truth.

\subsection{Why stage-1 reads are not rescored}

The decision to materialize the generic bank is a function of stage-1 data.
Rescoring the newly selected bank on those same observations would use the data
both for model selection and confirmation, invalidating the fixed-bank
conditional guarantee unless a selective-inference correction were supplied.
The protocol instead propagates the bank through the executed actions and starts
a new verifier on fresh data.

\subsection{Exact direct-shell certificate}

At a partial compiler node, let the current score components be
$(m,s_o,s_y)$, and let $r_o,r_y$ be the remaining opportunities to reduce the two
window scores. The lower bound
\begin{equation}
\begin{aligned}
L(u)=\bigl(&m,[s_o-r_o]_++[s_y-r_y]_+,\\[-0.2em]
&\max\{[s_o-r_o]_+,[s_y-r_y]_+\}\bigr)
\end{aligned}
\end{equation}
is admissible and lexicographically nondecreasing along every search path. If the
current endpoint-best map contains eight distinct shells and every open node has
lower bound strictly larger than the eighth shell, no unseen completion can
change any retained endpoint score. An arbitrary probe may propose an achievable
threshold, but only the exhaustive bounded search below that threshold supplies
the certificate. Failure invokes exact full enumeration.

\section{Additional empirical results}
\label{app:extra}

\subsection{Parallel speculation sensitivity}

A separate-worker sensitivity study compares deferred compilation, full
background compilation, evidence-gated background compilation, and actual
readwise switching on 72 new prospective streams. Evidence-gated background
compilation uses the same trigger as readwise switching but continues the nominal
verifier, spending 85 additional reads in aggregate. Full speculation provides a
0.258\% wall-clock advantage at 1 ms/read and 0.085\% at 10 ms/read while using
1.95 and 5.36 times as much compilation CPU, respectively; at 100 ms/read,
readwise switching is faster. The measured aggregate crossing is 12.78 ms/read.
This is a deployment sensitivity for the frozen compiler and traces, not a
hardware-independent theorem.

\subsection{Development replay}

Before the independent 2,000-operation shell confirmation, the same $J=8$
representation is replayed on the inherited 72 controller tasks. It remains
correct on 72/72, reduces fallback candidates by 62.20\%, reduces model steps by
62.02\%, and saves 98 physical reads. This replay is developmental; the frozen
independent split is the primary evidence.

\end{document}